\documentclass[10pt,journal,compsoc]{IEEEtran}
\ifCLASSOPTIONcompsoc
  \usepackage[nocompress]{cite}
\else
  \usepackage{cite}
\fi
\ifCLASSINFOpdf
\else
\fi

\usepackage{soul}
\usepackage{url}
\usepackage[hidelinks]{hyperref}
\usepackage[utf8]{inputenc}
\usepackage[small]{caption}
\usepackage{graphicx}
\usepackage{subcaption}
\usepackage{amsmath,amsfonts}
\usepackage{booktabs}
\usepackage{algorithm}
\usepackage{algorithmic}
\usepackage{multirow}
\usepackage{placeins}
\usepackage{blindtext}
\usepackage{xcolor}
\usepackage{balance}

\newcommand{\softmax}{\mathrm{softmax}}
\newcommand*{\myproofname}{My proof}

\newcommand{\argmin}{\arg\!\min}

\begin{document}
%
\title{Co-embedding of Nodes and Edges\\ with Graph Neural Networks}
%
%
%
%

\author{Xiaodong~Jiang,
        Ronghang~Zhu,
        Pengsheng~Ji,
        and~Sheng~Li,~\IEEEmembership{Senior~Member,~IEEE}
\IEEEcompsocitemizethanks{\IEEEcompsocthanksitem Xiaodong Jiang is with the Department of Statistics and the Department of Computer Science, University of Georgia, Athens, GA, 30602. E-mail: xiaodong.stat@gmail.com\protect
\IEEEcompsocthanksitem Ronghang Zhu is with the Department of Computer Science, University of Georgia, Athens, GA, 30602. E-mail: ronghangzhu@uga.edu\protect
\IEEEcompsocthanksitem Pengsheng Ji is with the Department of Statistics, University of Georgia, Athens, GA, 30602. E-mail: psji@uga.edu\protect
\IEEEcompsocthanksitem Sheng Li is with the Department of Computer Science and the Institute for Artificial Intelligence, University of Georgia, Athens, GA, 30602. E-mail: sheng.li@uga.edu\protect
}
\thanks{This manuscript has been accepted by the IEEE Transactions on Pattern Analysis and Machine Intelligence. (Corresponding author: Sheng Li)}}

%
%

\markboth{IEEE Transactions on Pattern Analysis and Machine Intelligence,~Vol.~X, No.~X, 2020}%
{Jiang \MakeLowercase{\textit{et al.}}: Co-embedding of Nodes and Edges with Graph Neural Networks}
%



\IEEEtitleabstractindextext{%
\begin{abstract}
Graph, as an important data representation, is ubiquitous in many real world applications ranging from social network analysis to biology. How to correctly and effectively learn and extract information from graph is essential for a large number of machine learning tasks. Graph embedding is a way to transform and encode the data structure in high dimensional and non-Euclidean feature space to a low dimensional and structural space, which is easily exploited by other machine learning algorithms. We have witnessed a huge surge of such embedding methods, from statistical approaches to recent deep learning methods such as the graph convolutional networks (GCN). Deep learning approaches usually outperform the traditional methods in most graph learning benchmarks by building an end-to-end learning framework to optimize the loss function directly. However, most of the existing GCN methods can only perform convolution operations with node features, while ignoring the handy information in edge features, such as relations in knowledge graphs. To address this problem, we present \textit{CensNet}, \textbf{C}onvolution with \textbf{E}dge-\textbf{N}ode \textbf{S}witching graph neural network, for learning tasks in graph-structured data with both node and edge features. CensNet is a general graph embedding framework, which embeds both nodes and edges to a latent feature space. By using \textit{line graph} of the original undirected graph, the role of nodes and edges are switched, and two novel graph convolution operations are proposed for feature propagation. Experimental results on real-world academic citation networks and quantum chemistry graphs show that our approach achieves or matches the state-of-the-art performance in four graph learning tasks, including semi-supervised node classification, multi-task graph classification, graph regression, and link prediction.
\end{abstract}

\begin{IEEEkeywords}
Graph Embedding, Graph Neural Networks, Line Graph, Node Classification, Link Prediction
\end{IEEEkeywords}}

\maketitle

\IEEEdisplaynontitleabstractindextext

%
\IEEEpeerreviewmaketitle

\IEEEraisesectionheading{\section{Introduction}\label{sec:introduction}}

%
%
%
%
\IEEEPARstart{D}{eep} learning models like convolutional neural networks (CNN) have been remarkably successful in many domains~\cite{DLnature2015}, including computer vision, natural language processing, signal processing, healthcare \cite{Esteva2018AGT}, etc. In particular, CNN and its variants are capable of extracting multi-scale localized spatial features and producing highly expressive representations~\cite{KrizhevskyCNN,ResNet}. Instead of using elaborated and sophisticated feature engineering procedures that heavily rely on domain expertise, CNN based models are usually trained in an end-to-end fashion, and they can model and learn the high-order interactions among input features for specific tasks. The convolution operations in CNN are well defined on data with underlying Euclidean structures (e.g., images), but they cannot be directly generalized to non-Euclidean data, such as graphs and manifolds~\cite{bronstein2017geometric}.

Graph-structured data is ubiquitous, from social network platforms to citation and co-authorship relations, from protein-protein interactions to chemical molecules~\cite{li2013low,ji2016coauthorship,li2015learning,li2017multi,gu2018rare,shi2019skeleton,franceschi2019learning}. Graph, as a complex data structure, is very effective in describing the relationships (edges) of objects (nodes). Due to the expressive power and flexibility of graph-structured data, graph neural networks (GNN) have attracted increasing attention in recent years, which try to adapt the effective deep representation learning approaches from Euclidean to non-Euclidean domains~\cite{GNNReview2019}. The earliest GNN method might be traced back to the work in~\cite{GNN2009}, which extends the general neural networks to graph domain. Along this research direction, many other GNN models have been proposed recently, such as the ChebNet~\cite{DefferrardNIPS2016}, graph convolutional networks (GCN)~\cite{Kipf2016SemiSupervisedCW}, GraphSAGE~\cite{Hamilton2017InductiveRL}, and Lanczosnet~\cite{liao2018lanczosnet}. By leveraging the node adjacency matrix, these GNN models analogously define convolution operators on graphs in either spectral or spatial spaces and have obtained promising performance in tasks like node classification~\cite{GNNReview2019}. With a few notable exceptions~\cite{Monti_2017_CVPR,GAT2018graph,Schlichtkrull2018ModelingRD,gong2019exploiting,gu2019scene}, GNN methods mainly focus on obtaining effective node embeddings, but ignore the information associated with edges that can be beneficial to many tasks such as node or edge classification, link prediction, community detection, and regression.   

\begin{figure*}[t]
\centering
\includegraphics[width=0.68\linewidth]{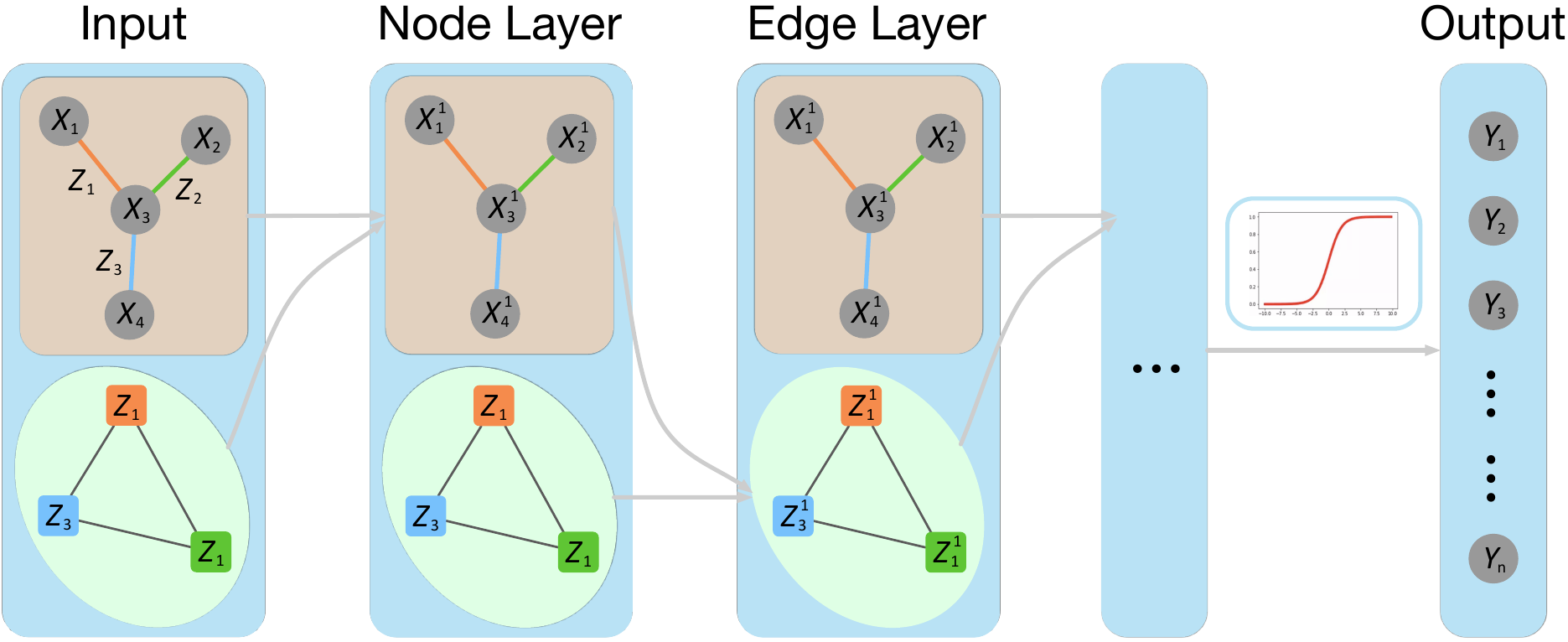}
\caption{Illustrative depiction of CensNet for node classification. The upper (orange color) components are convolution operations on node adjacency matrix and node features, while the lower (green color) components are the corresponding line graph convolution. Two types of layers as a combo, (1). Node Layer, update the node embedding with node and edge embedding from the previous layer, and (2). Edge Layer, update the edge embedding with the edge and node embedded features from the preceding layer.  The response is $n$-dimensional categorical variable where we assume there are $n$ nodes in the graph.}
\vspace{-0.0cm}
\label{figure_node_classification}
\end{figure*}

In this paper, we aim to learn both node embeddings and edge embeddings for graphs. Graphs in reality usually contain both node and edge features. In social networks, the node features could be demographic information or user behaviors, while the edge features might be the type of relationships or the years of friendship. In citation networks, the node features could be document-level embeddings of papers, and the edge features might indicate the common keywords or co-authors of two articles. In chemistry domain, each compound can be considered as a graph, where atoms are nodes, and properties of chemical bonds are edge features. We justify the motivation of jointly learning node and edge embeddings from the following two aspects. First, it is clear that edges and nodes always provide complementary feature information, which will be helpful for graph embedding. Second, learning edge embeddings is essential for edge-relevant tasks, such as edge classification and regression.    

Inspired by the \textit{Line Graph} in graph theory~\cite{lineGraph1960}, we propose a novel convolution with edge-node switching network (CensNet) for learning node and edge embeddings, as shown in Fig.~\ref{figure_node_classification}. Let $G$ denote the node adjacency matrix of a graph, its line graph $L(G)$ can be constructed to represent the adjacencies between edges of $G$. In our CensNet framework, the role of node and edge can be switched, and CensNet conducts the graph convolution operations on both the input graph $G$ and its line graph counterpart. With the help of node and edge features, CensNet employs two forward-pass feature propagation rules on $G$ and $L(G)$ to alternatively update the node and edge embeddings. Also, a mini-batch training algorithm for CensNet is devised to handle large-scale graphs.

This paper is a substantial extension of our previous work~\cite{Jiang2019CensNetCW}. Compared with~\cite{Jiang2019CensNetCW}, we extend CensNet to an unsupervised graph learning task, i.e., link prediction, by designing a novel network architecture and training algorithm. In addition, we have added a comprehensive review of related work, and provided detailed justifications of the proposed models, algorithms and experiments. The main contributions of this work are summarized as follows.
\begin{itemize}
\itemsep0em 
    \item \textbf{Co-embedding of nodes and edges.} The proposed CensNet is a general graph embedding framework, which can embed both nodes and edges to a latent feature space simultaneously, with the help of graph structure and node/edge features. Moreover, the node and edge embeddings can be mutually enriched owing to the alternative update rules.
    \item \textbf{Convolution on line graph.} Motivated by the line graph theory, we design novel convolution operations on both the input graph and its line graph counterpart, where the roles of node and edge are switched. 
    \item \textbf{Diverse learning tasks on graphs.} We apply CensNet to several graph-based learning tasks, including semi-supervised node classification, multi-task graph classification, graph regression, and unsupervised link prediction. CensNet can be customized for different tasks. For instance, the link prediction model is a hybrid structure of CensNet and Variational Autoencoder (VAE).
     
    \item \textbf{Extensive evaluations on benchmark data sets.} The extensive experiments on citation networks (Cora, Citeseer~\cite{citationGraph08}, and PubMed~\cite{pubmed12}) and Quantum Chemistry data sets (\textit{Tox21} and \textit{Lipophilicity}~\cite{MoleculeNet}) demonstrate that our method has achieved or matched state-of-the-art performance. 
\end{itemize}

The remaining part of the paper is organized as follows. Section \ref{related_works} reviews the related work in this domain and discusses the advantages of our method over existing methods. We provided the technical background and preliminaries in Section \ref{pre}. Section \ref{network_details} presents the main idea of CensNet, propagation rules and task dependent loss functions. Section \ref{applications} describes the applications of CensNet. To evaluate the model performance, we conduct extensive experiments of four graph learning tasks on five benchmark graph datasets, and present the results with discussions in Section \ref{exp}. Section~\ref{conclusion} concludes this paper with discussions and future work.

\section{Related Work}
\label{related_works}
In this section, we briefly introduce two related topics, including graph neural networks and learning on graphs. 

\subsection{Graph Neural Networks}
The Graph Neural Networks (GNN) model was first presented in \cite{GNN2009}, which extended existing traditional neural network models for processing the data represented in graph domains. Researchers then further proposed and simplified a set of graph-based neural network models for different graph learning tasks~\cite{morris2019weisfeiler,chen2019gated,liu2019geniepath,wu2019session}. These methods include but are not limited to the following. ChebyNet \cite{Defferrard2016ConvolutionalNN} is a GNN model with fast and localized spectral filtering. GCN \cite{Kipf2016SemiSupervisedCW} improves  efficiency with fast approximated spectral operations. GraphSAGE \cite{Hamilton2017InductiveRL} employs a general inductive framework that leverages node feature information to efficiently generate node embeddings for previously unseen data. Similarly, PinSAGE \cite{PinSAGE} is productionized for web-scale recommendation system at Pinterest. Other researchers then improved the graph convolution operations in many different ways. GAT \cite{GAT2018graph} introduces the attention mechanism to GNN, LNet \cite{liao2018lanczosnet} uses the Lanczos algorithm to construct low-rank approximations of the graph Laplacian for graph convolution. More recent methods include the graph Markov neural networks~\cite{qu2019gmnn}, position-aware graph neural networks~\cite{you2019position}, and conditional random field enhanced GCN~\cite{gao2019conditional}. We refer the readers to the comprehensive reviews for the recent algorithms and applications of GNN in~\cite{Zhou2018GraphNN} and~\cite{wu2019comprehensive}. 

Our CensNet approach is different from existing GNN methods in three aspects. First, CensNet can collectively learn both node and edge features in a graph, and thus it can be applied to edge-dependent tasks such as edge classification. Second, CensNet is flexible to use different graph kernels. We used approximated spectral kernel in this paper, while other kernels such as LNet \cite{liao2018lanczosnet} and SGC \cite{wu2019simplifying} can also be incorporated. Third, CensNet is capable of supervised, semi-supervised, and unsupervised learning tasks in graphs.

\subsection{Learning on Graphs}
Similar to classical machine learning tasks, graph learning also has different settings based on  data availability and real world needs. We briefly introduce the most commonly used (semi-)supervised and unsupervised graph learning tasks in this section. 

Supervised learning on graphs aims to train models with labeled data (e.g., labeled nodes), while semi-supervised learning trains models with both labeled and unlabeled data, especially when only a small number of labeled data is available. Due to the spatial connectivity property in graph, semi-supervised graph learning methods can leverage the information from unlabeled data. In the past decades, a number of sophisticated algorithms and models have been proposed, which can be categorized into two groups, regularized graph Laplacian based methods and graph embedding based methods \cite{Kipf2016SemiSupervisedCW}. The first group includes label propagation with Gaussian Fields and Harmonic Functions \cite{Zhu2003}, Manifold propagation \cite{Belkin2006}, and deep embeddings \cite{Weston2008}. The second group contains more recent works, such as DeepWalk \cite{Perozzi2014}, LINE \cite{Tang2015}, and node2vec \cite{Node2Vec}. These methods are inspired by the skip-gram model in \cite{Mikolov2013}, and they use various of random walk and search strategies. For most semi-supervised learning tasks, the aforementioned methods first learn the embeddings of graph, and then optimize the object functions with extra steps - the multi-step approaches cannot build an end-to-end optimization and learning framework. The recent advancements in deep learning can overcome this shortcoming and bridge the gap between graph embedding and specific learning tasks. Graph classification, graph regression, and node classification can be categorized to (semi)-supervised learning tasks, where the goal is to train a classifier or regression model to predict the unobserved labels/targets. 

On the other hand, unsupervised graph learning does not require any node/edge labels during training. The most representative unsupervised graph learning task might be link prediction, which predicts the existence of unobserved edges in a graph. Existing link prediction methods can be mainly categorized as four groups. (1) Similarity-based methods. Each pair of nodes is assigned a score with certain similarity metrics, to measure the probability of having an edge. Popular metrics including Local similarity \cite{liben2007link}, Global Similarity \cite{liben2005geographic}, and Quasi-Local Index \cite{lu2009similarity}. (2) Probabilistic and statistical methods with maximum likelihood estimation. These methods usually assume an underlying statistical models such as Stochastic Block Model \cite{holland1983stochastic} and Hierarchical Model \cite{clauset2008hierarchical}, and then estimate the corresponding maximum likelihood. (3) Algorithmic methods. This approach could formalize the link prediction task as a supervised classification problem, which can leverage the existing classification algorithms including Logistic regression, SVM, Tree-based methods, etc. Other algorithmic approaches can also refer to matrix factorization \cite{menon2011link} and low-rank approximation \cite{kunegis2009learning} methods. (4) Deep learning approaches. Recent research applies deep learning methods to link prediction, such as the Variational Autoencoder approach \cite{kipf2016variational}, GNN based method \cite{zhang2018link}, and neural bag-of-words model~\cite{kong2019neural}. 

In this paper, we apply the proposed CensNet to  supervised learning tasks (i.e., multi-task graph classification and graph regression), semi-supervised learning tasks (i.e., node classification) and unsupervised learning tasks (i.e., link prediction) on graphs. These applications fully demonstrate the flexibility and effectiveness of our CensNet.

\section{Preliminaries}
\label{pre}

\subsection{Notations}
\label{notations}

We present the mathematical notations for a graph with node and edge features in the following list.
\begin{itemize}
\item We assume an undirected graph $G$ 
with node set $V$ and edge set $E$, $E \subseteq V \times V$. $N_v = |V|$ and $N_e = |E|$ denote the number of nodes and edges, respectively. 
\item Let $A_{v} \in \mathbb{R}^{N_v \times N_v}$ be the adjacency matrix of $G$, where each element $A_v (i, j)$ denotes the connectivity of node $i$ and node $j$, where $i, j \in \{1, 2, ..., N_v\}$. $A_v$ is a binary matrix for an unweighted graph.
\item Let $X\in \mathbb{R}^{N_v \times d_v}$ be the node feature matrix, where each node is associated with a $d_v$-dimensional feature vector. 
\item Let $A_e \in \mathbb{R}^{N_e \times N_e}$ be the binary edge adjacency matrix of $G$, or node adjacency matrix of $L(G)$. $A_e (m, n)$ equals to 1 if the edge $m$ and edge $n$ are connected by a node in $G$, otherwise 0, where $m, n \in \{1, 2, ..., N_e\}$.
\item Let $Z \in \mathbb{R}^{N_e \times d_e}$ denote the edge feature matrix, where each edge has a $d_e$-dimensional feature vector.
\item Let $H_v^{(l)}$ be the $l$-th hidden layer of node convolution with $H_v^{(0)} = X$; Define $H_e^{(l)}$ as the $l$-th layer of edge convolution, and $H_e^{(0)} = Z$. 
\end{itemize}

In addition, we will also introduce other notations in the rest of the paper when necessary.

\subsection{Graph Convolution and Embedding}

Given a graph $G$ and its corresponding node feature matrix $X$, there are two major graph convolutions in literature, \textit{spectral} convolution in the Fourier domain and \textit{spatial} convolution in the node (or vertex) domain. We discard the subscripts of our notations for a moment, assuming $X$ is the node feature matrix, and $A$ is the node adjacency matrix.

\subsubsection{Spectral graph convolution} 

A spectral graph convolution is defined as the multiplication of a signal with a filter in the Fourier domain of the graph. The graph Fourier transform $Y$ is defined as the multiplication of a graph signal (e.g., node features $X$) with the eigenvector matrix $U$ of the graph Laplacian $L$, i.e., $Y=U^T X$ and $U\Sigma U^T=L$. The graph Laplacian $L$ can be defined in different ways: the simple Laplacian $D-A$, the symmetric normalized Laplacian $I-D^{- \frac{1}{2}}AD^{-\frac{1}{2}}$, or the random walk Laplacian $I-D^{-1}A$, where $I$ is an identity matrix and $D$ is the diagonal degree matrix. The symmetric normalized Laplacian is usually desirable due to the nice properties including symmetric, positive semidefinite, and all eigenvalues are in $[0,2]$. The ChebyNet \cite{DefferrardNIPS2016} and GCN \cite{Kipf2016SemiSupervisedCW} are based on spectral graph convolutions. The GCN proposes a layer-wise propagation rule based on an approximated graph spectral kernel as follows:
$$H^{(l+1)} = \sigma(\tilde{D}^{-\frac{1}{2}} \tilde{A} \tilde{D}^{-\frac{1}{2}}  H^{(l)} W^{(l)}),$$
where $\tilde{A}=A+I$ is the adjacency matrix with self-connections, $\tilde{D}$ is the degree matrix, $H^{(l)}$ and $W^{(l)}$ are the hidden feature matrix and learnable weight in the $l$-th layer.

\subsubsection{Spatial graph convolution} 

A spatial graph convolution is defined on the node domain, which can integrate or aggregate the signals among its neighbor nodes. MoNet~\cite{MoleculeNet} and GraphSAGE~\cite{Hamilton2017InductiveRL} are aggregation based representation learning models in this direction. MoNet is a generic spatial domain framework for deep learning on non-Euclidean domains such as graphs and manifolds. It adopts a spatial approach to learn different weights to a node’s neighbors. The node pseudo-coordinates was introduced to determine the relative position between a node and its neighbor. A weight function then map such relative position to the relative weights between the pairs of nodes. So the parameters of a graph filter are shared across different locations in the graph or manifold. GraphSAGE is another representative algorithm to learn node embeddings with spatial graph convolution, where the key idea is that it learns how to aggregate feature information from anode’s local neighborhood. It also adopts sampling to have a fixed number of neighbors for each node, and proposed a batch-training algorithm to save the memory and improve the training efficiency dramatically. 

\subsubsection{Node and Edge Embeddings}  

There have only been a few scattered examples that can link both node and edge features in a graph convolution simultaneously, which can be categorized as two approaches. The first one is to define different weight matrices for each relation or dimension on the edge features and aggregate different relation propagation in an additive fashion. In \cite{Schlichtkrull2018ModelingRD}, the following rule is used for the forward-pass update:
$$
h_i^{l+1} = \sigma (\sum_{r \in \mathcal{R} }^{} \sum_{j \in \mathcal{N}_i^{r}}^{} \frac{1}{c_{i,r}}  W_r^{(l)} h_j^{(l)} + W_0^{(l)} h_i^{(l)}),
$$  
where each relation $r \in \mathcal{R}$ has its corresponding weight matrix $W_r$. Such a simple aggregation enjoys the efficiency in computation but cannot capture the dependence of different relations and the interaction between the node and edge features. The second approach is to add the attention mechanism to graph convolutions by specifying different weights to different nodes in a neighborhood. MonNet~\cite{Monti_2017_CVPR} and GAT~\cite{GAT2018graph} are examples along this line. More importantly, both of these two approaches can only handle discrete or low-dimensional edge features.

Different from existing graph convolution and embedding methods, the proposed CensNet framework is independent of the choice of graph convolution, although we only implement one representative spectral-based method~\cite{Kipf2016SemiSupervisedCW} in this paper. Moreover, the co-embedding of nodes and edges in CensNet is inspired by \textit{line graph} and driven by mutually-enriched feature propagation rules, which can handle discrete or continuous node/edge features.

\section{Convolution with Edge-Node Switching}
\label{network_details}
\label{method}

In this section, we present the details of CensNet framework, derive two propagation rules for node and edge embeddings, and also introduce the task-dependent loss functions.

\subsection{Framework}
The CensNet framework consists of two types of layers, node layer and edge layer. Figure~\ref{figure_node_classification} shows the CensNet architecture for semi-supervised node classification. The input layer comprises of a node adjacency matrix and the corresponding node features, as well as its line graph counterpart - the edge adjacency matrix and edge features. The three colored edges (i.e., $z_1$, $z_2$ and $z_3$) in the sample graph are converted to three line graph nodes (squared shape). 

We define a CensNet combo as two types of layers, node layer and edge layer. In the node layer, all input data are processed to update the node embedding, while keeping the line graph (edge adjacency matrix and edge features) flow forward without any change. In the edge layer, we combine the updated node embedding with the line graph to update the edge embedding. Depending on the specific task and the availability of labels, CensNet adopts different types of activation functions for the node or edge embedding matrices. For example, in the graph node semi-classification task (e.g., paper classification in citation networks), we have the label for each node, and thus we can use the sigmoid function for the node embedding matrix in the final layer. For the graph classification or regression task (e.g., the molecular property prediction), we may apply an average pooling layer to reshape the node embeddings and obtain graph-level embeddings.  

We should note, the CensNet framework is a high-level abstract of the interactive graph embedding with both nodes and edges. One does not necessarily use the approximated spectral graph kernel for the convolution. Other techniques such as the kernels in Lanczos Network~\cite{liao2018lanczosnet} can also be applied to our framework.

\begin{figure*}[t!]
\centering
\includegraphics[width=0.8\linewidth]{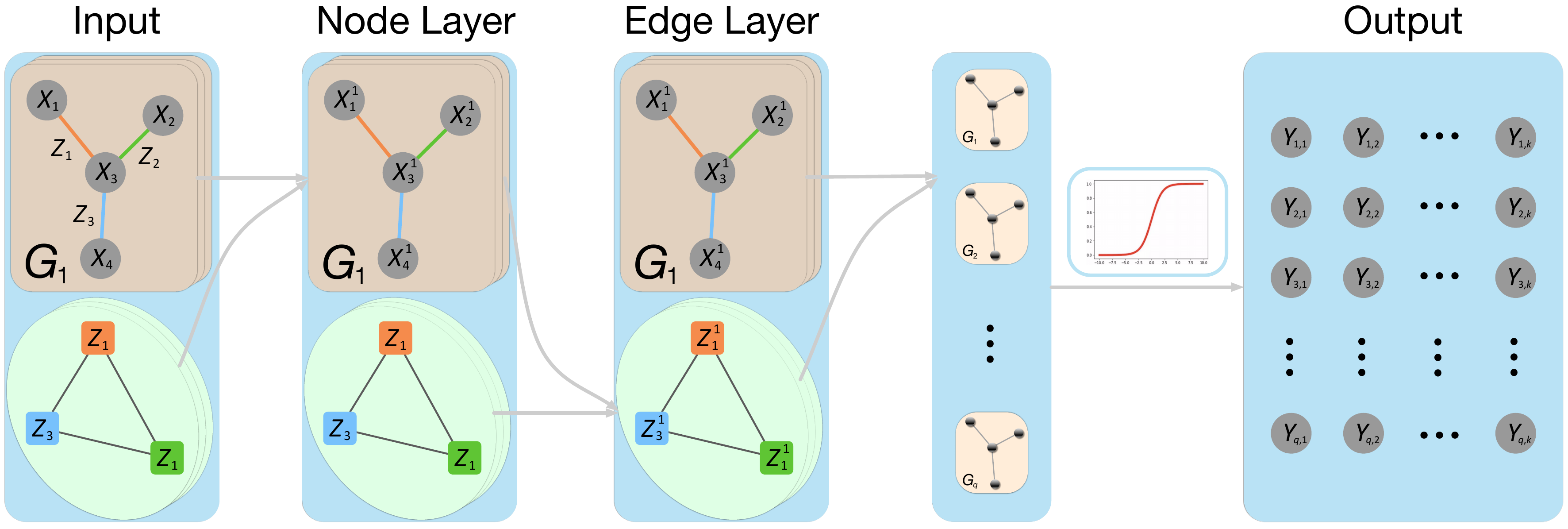}
\caption{Illustrative depiction of CensNet for graph classification and regression. Assume there are $q$ graphs, the upper (orange color) components are convolution operations on $q$ node adjacency matrix and $q$ node features, while the lower (green color) components are the corresponding $q$ line graph convolutions. The response is multi-dimensional categorical variables for multi-task graph classification, univariate continuous variable for graph regression. }
\label{figure_graph_tasks}
\end{figure*}

\subsection{Propagation Rules}
The CensNet uses approximated spectral graph convolution in the layer-wise propagation. We define the normalized (Laplacianized) node adjacency matrix with self-loop as:  
\begin{equation}
    \tilde{A}_v = D_v^{-\frac{1}{2}} (A_v + I_{N_v})D_v^{-\frac{1}{2} },
\end{equation}
where $D_v$ is the diagonal degree matrix of $A_v + I_{N_v}$, and $I$ is an identity matrix. 

\subsubsection{Propagation Rule for Node Layer} 
The layer-wise propagation rule for node feature in the $(l+1)$-th layer is defined as:
\begin{equation}
\label{node_prop}
    H_v^{(l+1)} = \sigma(T \Phi (H_e^{(l)} P_e) T^T \odot \tilde{A}_v H_v^{(l)} W_v), 
\end{equation}
where the matrix $T \in \mathbb{R}^{N_v \times N_e}$ is a binary transformation matrix and $T_{i, m}$ represents whether edge $m$ connects node $i$. Given the fact that each edge is formed by two nodes, each column of matrix $T$ has two elements being 1 and all others are 0. $P_e$ is a $d_e$-dimensional vector, defined as the learnable weights for edge feature vectors. $\Phi$ denotes the diagonalization operation, which places a one-dimensional vector to the diagonals of a square matrix. $\odot$ denotes the Hadamard product or element-wise product. Another view of this rule is to map corresponding elements from $T \Phi (H_e^{(l)} P_e) T^T$ to the normalized node adjacency matrix, and the $T \Phi (H_e^{(l)} P_e) T^T \odot \tilde{A}_v$ is a fused node adjacency matrix by using information from the line graph counterpart. The line graph brings zero impact if there is no physical edge, and thus we maintain the sparsity of the original graph, bringing significant computing benefits.

\subsubsection{Propagation Rule for Edge Layer} Similarly, the normalized (Laplacianized) edge adjacency matrix is defined as:
\begin{equation}
\tilde{A}_e = D_e^{-\frac{1}{2}} (A_e + I_{N_e})D_e^{-\frac{1}{2} },
\end{equation}
where $D_e$ is the degree matrix of  $A_e + I_{N_e}$. Furthermore, we define the propagation rule for edge features as follows:
\begin{equation}
\label{edge_prop}
    H_e^{(l+1)} = \sigma(T^T \Phi(H_v^{(l)} P_v) T \odot \tilde{A_e} H_e^{(l)} W_e).
\end{equation}
The $T$ matrix is the same as Equation \eqref{node_prop}, while $P_v$ represents the learnable weight for the nodes. As a return, the node feature and adjacency matrix are used to improve the edge embedding.
These two components bridge signals on nodes and edges, and the node and edge embeddings are updated alternatively.

\subsection{Task-Dependent Loss Functions}

The designs of the output layer, as well as the loss function, are task-dependent. For node or edge classification tasks, we may apply the sigmoid function to the final hidden node or edge layers; for the graph classification task, we need an extra pooling operation that maps the node-level embeddings to a graph-level representation. For example, we take the average prediction of the atoms in a molecule as the graph-level output. The graph-level max pooling might not be appropriate from a practical point of view; a learnable parameter (weight) for each node is also not acceptable because of the tendency to overfitting. In the next section, we will describe how to design loss functions based on CensNet for different tasks, including node classification, multi-task graph classification, graph regression, and link prediction.

\section{Applications of CensNet}
\label{applications}
We demonstrate that CensNet is flexible and powerful to learn the graph embeddings for four graph learning tasks, including semi-supervised node classification, multi-task graph classification, graph regression, and link prediction. We define these tasks, network structures and training algorithms in following sub-sections, and present the corresponding experimental results in Section \ref{exp}.

\begin{figure*}[t!]
\centering
\includegraphics[width=0.8\linewidth]{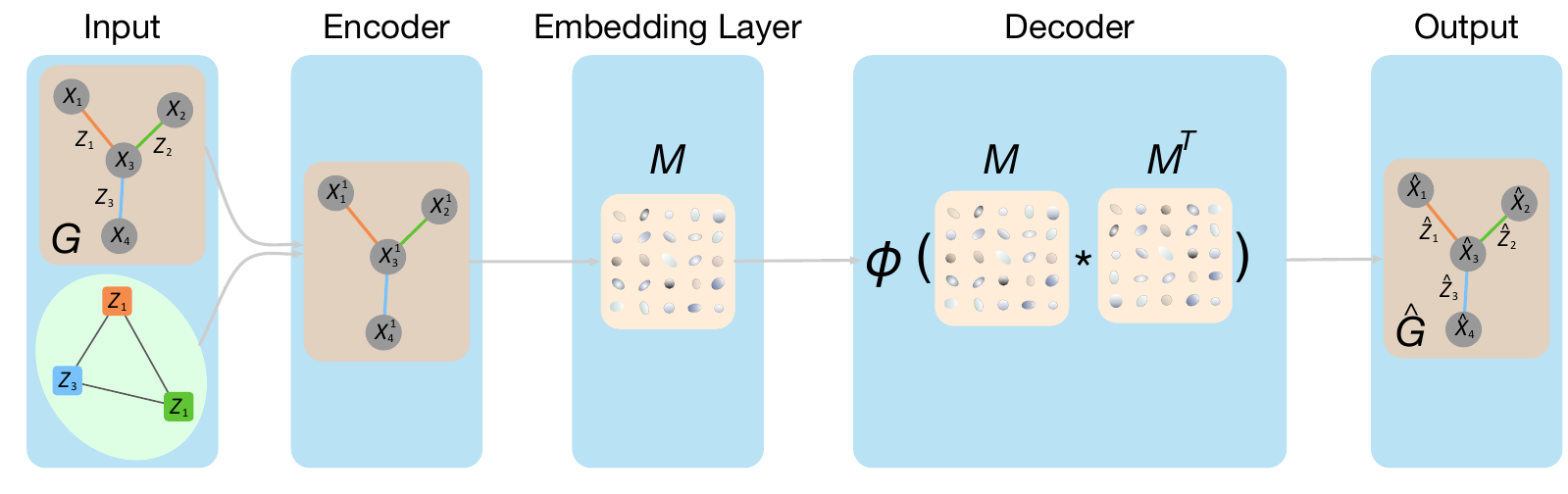}
\caption{Illustration of CensNet-VAE for unsupervised link prediction. We use the edge adjacency matrix to update the node feature embedding, which is the input for the encoder, and then pass the embedding layer generated from encoder to decoder. Finally, embeddings are used to reconstruct the graph.}
\label{figure_cens_vae}
\end{figure*}

\subsection{Semi-Supervised Node Classification}
Semi-supervised node classification is an important task in deep graph learning, which might be the first domain that graph convolutional networks dramatically outperformed classical methods. The goal is to classify nodes into different categories in a graph (e.g., different research areas in a citation network), where the labels are only available for a small subset of nodes. Figure \ref{figure_node_classification} shows the basic blocks of CensNet - the node layer updates the node embedding with node and edge embedding from the previous layer, and the edge layer updates the edge embedding with the edge and node embedded features from the preceding layer. The response is $n$-dimensional categorical variable where we assume there are $n$ nodes in the graph. 

For the semi-supervised node classification task, the loss function is defined as:
\begin{equation} 
\label{loss_node_classification}
\mathcal{L}(\Theta) = -\sum_{l\in Y_L}^{} \sum_{f=1}^{F} Y_{lf} \log{M_{lf}},
\end{equation}
where $Y_L$ is a set of nodes with labels, and $M$ is the softmax results of the output from the last node layer of CensNet, assuming the final node feature map has $F$ dimensions. The $\Theta = (W_v, W_e, P_v, P_e)$ is the parameter set, which is also used in other three tasks. Please note, we use the same notations interchangeably for node-level and graph-level classification tasks as an illustration purpose. 

\subsection{Multi-Task Graph Classification}
Multi-task graph classification is a problem with practical applications in many different domains. The goal is to classify a set of (usually small-sized) graphs to multiple labels. Chemists and Biologists are usually interested in toxicological outcomes of certain compound from their experiments, and these outcomes can be denoted as multi-labels, while the compounds in experiments are represented as graphs with atoms as nodes and chemical bonds as edges in each graph. 

Figure \ref{figure_graph_tasks} illustrates the depiction of CensNet for graph classification. Let's assume there are $q$ graphs and each graph has $k$ tasks ($k$-dimensional target labels) and our goal is to predict the $q \times k$ target matrix. Similar to Figure \ref{figure_node_classification}, we combine the node adjacency matrix and it's corresponding edge adjacency matrix to update the node layer, and sum the single-graph loss function in Equation \ref{loss_node_classification} from $q$ graphs as the final loss function. Please note, for implementation, we may stack the $q$ node adjacency matrix as a giant node adjacency matrix where each sub-adjacency is placed in diagonal, or we can implement the batch-training algorithm to train with a small number of graphs in each batch. 

Specifically, the loss function in this task is defined as follows:
\begin{equation} 
\mathcal{L}(\Theta) = -\sum_{i=1}^{q}(\sum_{l\in Y_L}^{k} \sum_{f=1}^{F} Y_{lf} \log{M_{lf}})_i,
\end{equation}
where, as we mentioned, the loss function is the sum of cross entropy from $k$ tasks. $Y_L$ is a set of graphs with labels, and $M$ is the softmax results of the output from the last layer, assuming the final feature map has $F$ dimensions. 

\subsection{Graph Regression}
Similar to graph classification, researchers might also be interested in continuous and numeric measures for chemical compounds, which can be formalized as a regression problem with graph level response.   

For the regression task where the response is a continuous variable, we define the following loss function by using the $l^p$ regularized mean square error (MSE):
\begin{equation}
\mathcal{L}(\Theta) = \sum_{l\in Y_L}^{} \sum_{f=1}^{F} ||Y_{lf} - \hat{Y}_{lf}||_2^2  + \lambda ||\Theta||_p,
\end{equation}
where $\hat{Y}$ is the predicted outcome from the last node hidden layer of CensNet. The regularized term is used to control  model complexity and avoid overfitting.

\subsection{Link Prediction}
Link prediction aims to predict the existence of edges given the node features and existing edges and their associated features.
It is different from previous tasks, as the node label information is usually not considered. We designed an encoder-decoder network structure for link prediction, which combines CensNet with Variational Autoencoder (VAE), denoted as CensNet-VAE. 

Following the variational graph auto-encoders~\cite{kipf2016variational}, we denote by $G$ an undirected graph with node feature matrix $X$, edge feature matrix $Z$, node adjacency matrix $A_v$, and node adjacency matrix with self-loop, $\tilde{A}_v$. The latent embedding matrix (i.e., Embedding Layer in Fig.~\ref{figure_cens_vae}) is denoted as $M$. Next, we introduce the details of the encoder model and decoder model of CensNet-VAE.

\textbf{Encoder model.} Inspired by~\cite{kipf2016variational,Hasanzadeh2019SIGVAE}, we use an encoder model parameterized by a two-layer network:
\begin{equation}
Q{(M|X, Z, \tilde{A}_v)} = \prod_{i=1}^{N_v}Q({m}_{i}|{X, Z, \tilde{A}_v}),
\end{equation}
where $Q({m}_{i}|{X, Z, \tilde{A}_v})=\mathcal{N}({m}_{i}|\boldsymbol{\mu}_{i},\text{diag}(\boldsymbol{\sigma}_{i}^{2}))$, $\boldsymbol{\mu} = \textbf{Encoder}_{\boldsymbol{\mu}}(X, Z, \tilde{A}_v)$ is the matrix of mean vectors $\boldsymbol{\mu}_i$ and $\log{\boldsymbol{\sigma}} = \textbf{Encoder}_{\boldsymbol{\sigma}}(X, Z, \tilde{A}_v)$. The two-layer network $\textbf{Encoder}$ is defined as $\textbf{Encoder}(X, Z, \tilde{A}_v) = \tilde{A}_v \text{ReLU} (\textbf{CensNet}_N(X, Z, \tilde{A}_v))W_i$, where $\textbf{CensNet}_N(X, Z, \tilde{A}_v)$ is the node layer which uses node layer propagation rule, and $W_i$ denote learnable weight matrices for $\boldsymbol{\mu}$ and $\boldsymbol{\sigma}$.

\textbf{Decoder model. }The decoder model is a non-probabilistic variant model defined as follows:
\begin{equation}
P(\tilde{A}_v|M) =\prod_{i=1}^{N}\prod_{j=1}^{N}P(\tilde{A}_{vij}|m_{i},m_{j}),
\end{equation}
where $P(\tilde{A}_{vij}=1|{m}_{i},{m}_{j})=\phi(m_{i}^{T},m_{i})$, ${\tilde{A}_v}{}_{ij}$ are the elements of $\tilde{A}_v$ and  $\phi(\cdot)$ is the logistic sigmoid function. We get the reconstructed node adjacency matrix $\hat{A}_v$ as
\begin{equation}
\hat{A}_v = \phi(MM^{\top}).
\end{equation}

By combing the encoder and decoder models, the loss function of CensNet-VAE is written as:
\begin{small}
\begin{equation}
\mathcal{L}(\Theta)=\mathbb{E}_{Q{(M|X, Z, \tilde{A}_v)}}[\log{P{(\tilde{A}_v|M)}}-\mathrm{KL}[Q{(M|X, Z, \tilde{A}_v)}||P{(M)]},
\label{eqlink}
\end{equation}
\end{small}
where $M$ is the embedding matrix, and $\mathrm{KL}[Q(\cdot)||P(\cdot)]$ denotes the Kullback-Leibler divergence between $Q(\cdot)$ and $P(\cdot)$. We use a Gaussian prior $P(M) = \prod_{i}P(m_{i})=\prod_{i}\mathcal{N}(m_{i}|0, \textbf{I}) $.

As shown in Fig.~\ref{figure_cens_vae}, we only employ one edge layer for CensNet-VAE, although multiple edge and node layers could be easily incorporated. The reason is that adding more layers will not always lead to significant performance gain, but will propagate noisy information and largely increase the model complexity. This observation is consistent with the analysis of graph convolutional networks in~\cite{Kipf2016SemiSupervisedCW}.

\begin{table*}[ht!]
\begin{center}
\caption{Node classification accuracy (in percent) on three citation graph data sets.}
\label{citation_results}
\begin{tabular}{|c|c|c|c|c|c|c|c|c|}
    \hline
    Data & TrainPCT & ChebyNet & GCN & GraphSAGE & GAT & LNet  & AdaLNet & CensNet (Ours)  \\\cline{1-9} 
    \multirow{3}{*}{Cora}&3\%   & 62.1$\pm$6.7 & 74.0$\pm$2.8 & 64.2$\pm$4.0 & 56.8$\pm$7.9 & 76.3$\pm$2.3 & 77.7$\pm$2.4 & \textbf{79.4$\pm$1.0}  \\
    &1\%   & 44.2$\pm$5.6 & 61.0$\pm$7.2 & 49.0$\pm$5.8 & 48.6$\pm$8.0 & 66.1$\pm$8.2 & \textbf{67.5$\pm$8.7} & 67.1$\pm$1.3  \\
    &0.5\% & 33.9$\pm$5.0 & 52.9$\pm$7.4 & 37.5$\pm$5.4 & 41.4$\pm$6.9 & 58.1$\pm$8.2 & \textbf{60.8$\pm$9.0} &  57.7$\pm$3.9  \\
    \hline
    \hline
    \multirow{3}{*}{Citeseer}&1\% & 59.4$\pm$5.4 & 58.3$\pm$4.0 & 51.0$\pm$5.7 & 46.5$\pm$9.3 & 61.3$\pm$3.9 & \textbf{63.3$\pm$1.8} & 62.5$\pm$1.5  \\
    &0.5\%   & 45.3$\pm$6.6 & 47.7$\pm$4.4 & 33.8$\pm$7.0 & 38.2$\pm$7.1 & 53.2$\pm$4.0 & 53.8$\pm$4.7 & \textbf{57.6$\pm$3.0}  \\
    &0.3\% & 39.3$\pm$4.9 & 39.2$\pm$6.3 & 25.7$\pm$6.1 & 30.9$\pm$6.9 & 44.4$\pm$4.5 & 46.7$\pm$5.6 &  \textbf{49.4$\pm$3.6}  \\
    \hline
    \hline
        \multirow{3}{*}{PubMed}&0.1\% & 55.2$\pm$6.8 & 73.0$\pm$5.5 & 65.4$\pm$6.2 & 59.6$\pm$9.5 & \textbf{73.4$\pm$5.1} & 72.8$\pm$4.6 & 69.9$\pm$2.1  \\
    &0.05\%   & 48.2$\pm$7.4 & 64.6$\pm$7.5 & 53.0$\pm$8.0 & 50.4$\pm$9.7 & \textbf{68.8$\pm$5.6} & 66.0$\pm$4.5 & 65.7$\pm$1.2  \\
    &0.03\% & 45.3$\pm$4.5 & 57.9$\pm$8.1 & 45.4$\pm$5.5 & 50.9$\pm$8.8 & 60.4$\pm$8.6 & 61.0$\pm$8.7 &  \textbf{61.4$\pm$2.8}  \\
\hline
\end{tabular}
\end{center}
\vspace{0mm}
\vspace{0mm}
\end{table*}

\begin{algorithm}[tb!]
   \caption{CensNet for Node Classification}
   \label{alg:CensNet}
\begin{algorithmic}[1]
   \STATE {\bfseries Input:}  
    \textit{Node adjacency matrix} $A_v$\\
    \textit{node feature matrix} $X$\\
    \textit{node label} $Y$\\ 
    \textit{edge feature matrix} $Z$\\
    \textit{nonlinearity} $\sigma$ 
   \STATE Build \textit{transformation matrix} $T$ and \textit{edge adjacency matrix} ${A_e}$ 
    \STATE Run Adam to minimize $\mathcal{L}$ 
   $$W_v, W_e, P_v, P_e:= \argmin_{W_v, W_e, P_v, P_e} \mathcal{L}$$ 
   \STATE {\bfseries Output:} 
   $\softmax(H_v)$ 
\end{algorithmic}
\end{algorithm}

\begin{algorithm}[t]
   \caption{Mini-batch CensNet}
   \label{alg:minibatch}
\begin{algorithmic}[1]
   \FOR{each epoch}
   \STATE Construct batches to make each batch containing subgraph $G' \in G$ and the nodes in $G'$ are proportionally selected from training, validation, and testing sets. 
   \FOR{each batch}  
        \STATE Run Algorithm \ref{alg:CensNet}
   \ENDFOR
   \ENDFOR
\end{algorithmic}
\end{algorithm}

\subsection{Training Algorithms}
We show our algorithm for semi-supervised node classification in \textit{Algorithm~\ref{alg:CensNet}}, which uses the layer-wise propagation rules defined in Equations \eqref{node_prop} and \eqref{edge_prop}. The algorithms for the other three tasks can be summarized in a similar way. Optimization algorithms, such as Adam or SGD, are employed to deal with the cross-entropy loss function. When the input data cannot fit into the GPU memory, a mini-batch strategy is usually preferred~\cite{adam}. Thus, we also design a mini-batch training algorithm for large graphs in \textit{Algorithm~\ref{alg:minibatch}} for the node classification task with few labeled nodes. The key idea is to sample the nodes from training, validation, and test set proportionally, to construct the batches. Our empirical results show that such a training strategy could match the performance of training with the entire graph.

\section{Experiments}
\label{exp}
In this section, we provide comprehensive evaluations of the proposed CensNet method and baselines in four graph learning tasks on five benchmark data sets.

\subsection{Semi-supervised Node Classification}

\subsubsection{Datasets and Settings} 

Three benchmark datasets are employed in our experiments for the semi-supervised node classification task, including Cora, Citeseer, and PubMed~\cite{giles1998citeseer,mccallum2000automating,citationGraph08}. These datasets has been analyzed by many graph convolutional network models such as the ones in \cite{DefferrardNIPS2016,Kipf2016SemiSupervisedCW,Hamilton2017InductiveRL,GAT2018graph,liao2018lanczosnet}. \textit{Cora} has 2,708 nodes (papers) and 5,429 edges (citation links), and each node has 1,433 tf-idf features. The papers are classified into 7 different research areas thus the response has seven different values. \textit{Citeseer} has 3,327 nodes and 4,732 edges with 3,703 node features; the papers are grouped into 6 research fields. \textit{PubMed} contains 19,717 nodes and 44,338 edges, each node has 500 features and the papers are in 3 categories. Please remark that these citation graphs are not naturally good benchmarks for CensNet because there is no available edge feature. However, we still run our algorithm with effortless hand-crafted edge features to show the competitive performance of CensNet. Examples of such edge features could be the pairwise node feature correlations or cosine similarities.

For three citation graphs, we create two simple edge features: (1) the pairwise cosine similarities between corresponding node features, and (2) a 2-dimensional vector to represent the edge directions. If paper A cites paper B then the vector is $[1,0]$, otherwise $[0,1]$.

\begin{table*}[ht!]
    \centering
    \caption{Experimental results on \textit{Tox21} and \textit{Lipophilicity} data sets.}
    \begin{tabular}{|c|c|c|c|c|c|c|c|c|c|}
    \hline
    Train&Data& \multicolumn{4}{|c|}{\textit{Tox21} (AUC)}&\multicolumn{4}{|c|}{ \textit{Lipophilicity} (RMSE)}\\
    \cline{3-10}
     PCT& Split & RF&Logistic&GCN&CensNet&RF&LR&GCN&CensNet\\
    \hline
    \multirow{2}{*}{60\%} &Val.& 0.69$\pm$0.01& 0.69$\pm$0.01 & 0.72$\pm$0.00 & \textbf{0.76$\pm$0.00} &1.19$\pm$0.01 & 1.46$\pm$0.37 & 1.04$\pm$0.01 & \textbf{0.94$\pm$0.01}\\
         &Test                 & 0.71$\pm$0.01& 0.71$\pm$0.01 & 0.73$\pm$0.00  &\textbf{0.77$\pm$0.00}&1.16$\pm$ 0.02 &1.17 $\pm$0.03 & 1.06$\pm$0.00&\textbf{0.97$\pm$0.01}\\ 
    \hline
    \hline
    \multirow{2}{*}{70\%} &Val. & 0.70 $\pm$ 0.01 &0.70$\pm$0.01 & 0.73$\pm$0.00& \textbf{0.76$\pm$0.00}  & 1.18$\pm$0.02 & 1.19$\pm$0.01 & 1.02$\pm$0.01 & \textbf{0.92 $\pm$0.01}\\
                          &Test & 0.70 $\pm$ 0.01  &0.71$\pm$0.01  &  0.74$\pm$0.00 & \textbf{0.77$\pm$0.00} &1.16$\pm$0.02 & 1.17$\pm$0.04& 1.05 $\pm$0.01&  \textbf{0.95$\pm$0.01}\\ 
    \hline
    \hline
         \multirow{2}{*}{80\%} &Val.& 0.71$\pm$0.01 & 0.71$\pm$0.01 & 0.72$\pm$0.00& \textbf{0.76$\pm$0.00} & 1.17$\pm$0.02 & 1.16$\pm$0.02 & 1.05$\pm$0.01  & \textbf{0.96$\pm$0.01} \\
          &Test &0.71$\pm$0.01 & 0.71$\pm$0.01&  0.75$\pm$0.00& \textbf{0.78$\pm$0.00} &1.16$\pm$0.01  & 1.15$\pm$0.01 &  1.05$\pm$0.01  & \textbf{0.93$\pm$0.01}\\ 
    \hline
    \hline
\multirow{2}{*}{90\%} &Val. & 0.71$\pm$0.02 & 0.71$\pm$0.01 & 0.74$\pm$0.00 & \textbf{0.78$\pm$0.01} & 1.18$\pm$0.02 & 1.18$\pm$0.03 & 1.08$\pm$0.00 & \textbf{0.94$\pm$0.02} \\
                      &Test & 0.71$\pm$0.02 & 0.71$\pm$0.02  &  0.75$\pm$0.00 & \textbf{0.79$\pm$0.01}& 1.13$\pm$0.03 & 1.13$\pm$0.03  & 0.97$\pm$0.00 & \textbf{0.83$\pm$0.02}  \\ 
    \hline
    \end{tabular}
    \vspace{0mm}
    \label{chem_result}
    \vspace{0mm}
\end{table*}

Semi-supervised node classification is a classical task in the graph learning and statistical learning communities. We adopt a few-shot learning strategy, i.e., only keep a small number of labeled data in training set while splitting the rest of the data for validation and test. We follow the splitting strategy in \cite{liao2018lanczosnet} and implement the experiments for different label rate. We evaluate our method and baselines with 3\%, 1\% and 0.5\% labeled data in training set on \textit{Cora}, 1\%, 0.5\% and 0.3\% on \textit{Citeseer}, and 0.1\%, 0.05\% and 0.03\% on \textit{PubMed}. For all three data sets, we randomly select 50\% for validation and the rest for testing. We compare CensNet with seven representative graph convolution networks when using different percentages of labeled data. The seven baselines include ChebyNet~\cite{Defferrard2016ConvolutionalNN}, GCN~\cite{Kipf2016SemiSupervisedCW}, GraphSAGE~\cite{Hamilton2017InductiveRL}, GAT~\cite{GAT2018graph}, LNet and AdaLet~\cite{liao2018lanczosnet}. To the best of our knowledge, LNet and AdaLet are the state-of-the-art methods for this task~\cite{liao2018lanczosnet}, and we will present a comprehensive comparison between these two methods and the proposed CensNet.

All experiments are conducted on an Azure Linux VM (CPU: Intel(R) Xeon(R) CPU E5-2690 v3, GPU: NVIDIA Tesla K80). We implemented all graph convolution network algorithms in PyTorch~\cite{paszke2017automatic} v1.0.0. For other classical algorithms (random forest, linear regression, logistic regression), we used the implementations in the Python package Scikit-learn~\cite{scikit-learn}. For graph convolution models, we didn't implement any sophisticated fine-tuning strategies but tried different settings of learning rate from $\{0.01, 0.005, 0.001, 0.0005\}$, batch size $\{16, 32, 64, 128, 256\}$, number of epochs $\{200, 300, 500, 1000\}$, etc. We implemented three layers architecture (Node Layer - Edge Layer - Node Layer) with 32 units in each hidden layer. We report the best-performed results for each algorithm. For tree-based methods, we use cross-validation to tune the parameters; for linear and logistic regression models, we run the algorithms without using variable selection.

\subsubsection{Results and Discussions} 

We follow the same experimental settings in~\cite{liao2018lanczosnet} and re-use the benchmark results as our baselines. Table~\ref{citation_results} shows the node classification results of CensNet and other baselines. For all the experiments, we observed significant overfitting with low label rates for all graph convolution networks. We highlight the best-performed method (with the highest accuracy in the test set) in each setting for all three data sets. Our CensNet method obtains the best accuracy in 4 out of 9 experiments, which is followed by LNet and AdaLNet - the state-of-the-art algorithms for this task. One may believe that adding edge information to the algorithm is not a fair comparison to the benchmarks; however, our newly created edge features are all from the benchmark data; thus no extra signal is introduced. The classical GCN~\cite{Kipf2016SemiSupervisedCW} also achieves competitive results in most scenarios, which coincides the conclusion from the extensive experiments in~\cite{GCNPitfalls2018}.

To evaluate the efficiency of our method, we compare the computing time of our method and GCN. The node-edge switching operation in our method introduces extra computational cost. For example, CensNet is about 40\% slower than GCN per epoch for semi-supervised node classification on the Cora dataset.

\begin{figure}[t]
\centering
\includegraphics[width=0.9\linewidth]{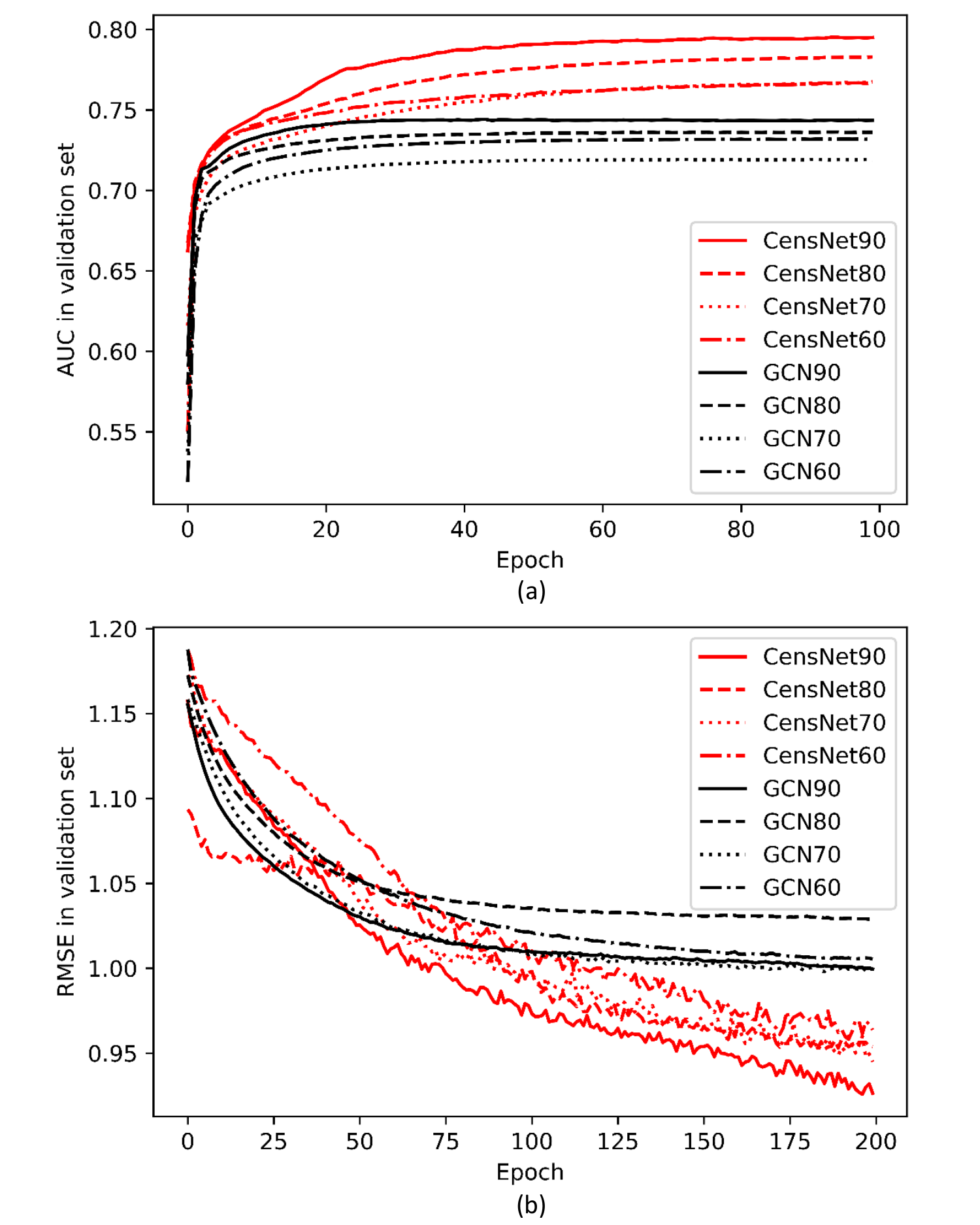}
\vspace{0in}
\caption{AUC/RMSE in validation set for \textit{Tox21/Lipophilicity}. The name of each curve is formed with algorithm name and label ratio in training set. For example, CensNet90 means the CensNet algorithm with 90\% data in training set.}
\vspace{0in}
\label{figure2}
\end{figure}

\begin{table*}[h!]
\begin{center}
\caption{Link prediction AUC and AP (in percent) on three citation graph data sets.}
\label{linkprediction_results}
\begin{tabular}{|c|c|c|c|c|c|c|}
\hline
\multicolumn{1}{|c|}{ \multirow{2}*{Method} } &\multicolumn{2}{c|}{Cora} &\multicolumn{2}{c|}{Citeseer} &\multicolumn{2}{c|}{Pubmed} \\
\cline{2-7}
\multicolumn{1}{|c|}{}&AUC&AP&AUC&AP&AUC&AP\\
\hline
SC & 84.6$\pm$0.01 & 88.5$\pm$0.00 & 80.5$\pm$0.01 & 85.0$\pm$0.01 & 84.2$\pm$0.02 & 87.7$\pm$0.01 \\
DW & 83.1$\pm$0.01 & 85.0$\pm$0.00 & 80.5$\pm$0.02 & 83.6$\pm$0.01 & 84.2$\pm$0.00 & 84.1$\pm$0.00 \\
VGAE & 91.4$\pm$0.01 & 92.6$\pm$0.01 & \textbf{90.8$\pm$0.02} & \textbf{92.0$\pm$0.02} & 94.4$\pm$0.02 & 94.7$\pm$0.02 \\ 
CensNet-VAE & \textbf{91.7$\pm$0.02} & \textbf{92.6$\pm$0.01} & 90.6$\pm$0.01 & 91.6$\pm$0.01 & \textbf{95.5$\pm$0.03} & \textbf{95.9$\pm$0.02} \\
\hline
\end{tabular}
\end{center}
\vspace{0mm}
\vspace{0mm}
\end{table*}

\subsection{Multi-Task Graph Classification}\label{graphClass}

\subsubsection{Dataset and Settings} 

The \textit{Toxicology in the 21st Century (Tox21, \cite{MoleculeNet})} initiative created a public database measuring the toxicity of compounds, which has been used in the 2014 \textit{Tox21} Data Challenge. This dataset contains quantitative toxicity measurements for 7,831 environmental compounds and drugs. A compound structure (in SMILE format) is usually expressed as a graph with atoms as nodes and bonds being edges. There are 55 bond features, and each atom has 25 features. Each compound is associated with 12 binary labels that represent the outcome (active/inactive) of 12 different toxicological experiments. There are about 20\% missing values in these labels, and we exclude those observations when computing the loss but still keep them in the training process. In general, the \textit{Tox21} data is designed for multi-task graph-level classification with multi-dimensional node and edge features.

We follow the preprocessing steps in~\cite{MoleculeNet} to convert each compound in \textit{Tox21} to a small graph, and remove the compounds whose SMILE representation cannot be converted to a graph structure. We then randomly split the dataset to different partitions as training, validation and test sets, respectively. We consider 4 data splitting settings, by keeping 60\%, 70\%, 80\%, and 90\% of the molecule graphs as the training set, while equally breaking the rest of the data sets as validation and test sets.

We evaluate the performance of CensNet on the \textit{Tox21} dataset under 4 data splitting scenarios. In \cite{MoleculeNet}, a comprehensive model comparisons on the \textit{Tox21} data is presented, and the state-of-the-art method is GCN~\cite{Kipf2016SemiSupervisedCW}. We implemented the mini-batch GCN, and other two classical classification algorithms, logistic regression and random forest, as three baselines. We report the area under the ROC curve (AUC) metric in both validation and test data sets for all compared methods.

\subsubsection{Results and Discussions} 

Table~\ref{chem_result} shows the AUC values on the validation and test sets, with four different training label rates. We replicate all experiments three times and report the mean and standard deviation of AUC values. The CensNet algorithm maintains significant advantages over all other methods in all settings, while both GCN and CensNet perform better than the other two traditional methods. The logistic regression and random forest models can hardly capture the association between signals and response, even with increased training sets. Figure~\ref{figure2}(a) shows the changes in validation AUC in 200 epochs. The GCN's curve becomes flat within 20 epochs while the CensNet can continuously improve. Also, the CensNet with smaller training sets can beat the GCN with more training data, uniformly and consistently.


\subsection{Graph Regression}

\subsubsection{Dataset and Settings}

The \textit{Lipophilicity} is an important feature of drug molecules that affects both membrane permeability and solubility \cite{MoleculeNet}. This dataset provides experimental results of octanol/water distribution coefficient (logD at pH 7.4) of 4,200 compounds. There are 34 features on the edge (bond) and 25 features on the node (atom), while the response is one single continuous variable.

We follow the same preprocessing as described in Section~\ref{graphClass}. We randomly split the dataset to different partitions as training, validation and test sets, respectively. We consider 4 data splitting settings, by keeping 60\%, 70\%, 80\%, and 90\% of the molecule graphs as the training set, while equally breaking the rest of the data sets as validation and test sets.

The graph regression task is similar to graph classification but using different loss function and performance metric. We use the same data splitting strategies as in \textit{Tox21}. The baseline algorithms are GCN, linear regression, and random forest regression. We also report the root mean square error (RMSE) in both validation and test sets.

\subsubsection{Results and Discussions} 

Compared with three baseline methods, our CensNet has achieved the best performance in both validation and test sets under four training settings. For a fair comparison, we replicate all experiments three times and report the mean RMSE with standard errors. The following conclusions can be summarized from Table~\ref{chem_result}: (1). All these 4 algorithms achieve better performance with larger training sets, while random forest and logistic regression only obtain limited performance lift. (2). Both GCN and CensNet have large positive margins over traditional methods, indicating that the graph structure is not neglectable in the molecule regression task. (3). CensNet improves the performance of GCN by 5\% - 15\% in RMSE, which implies that considering the edge features in the molecule can improve the quality of node embedding, as a consequence, bring significant benefits to the learning process. Figure \ref{figure2}(b) shows that the CensNet gains leading positions after around 50-75 epochs, while GCN's curves still keep flat.

\subsection{Link Prediction}

\subsubsection{Datasets and Settings} 

For link prediction, we benchmarked the three citation graphs with the same edge features for semi-supservised node classification, i.e., Cora, Citeseer and Pubmed. We follow the same experimental setting in~\cite{kipf2016variational} that 85\% of citation links/edges are used as train sets, 10\% and 5\% of citation links as positive instances for testing and validating, and use the same number of randomly sampled pairs of unconnected nodes (fake-edges) as negative samples for test and validation sets. The area under the ROC-curve (AUC) and average precision (AP) are two evaluation metrics in this experiment.

For CensNet-VAE, we initialize weights as described in~\cite{XavierInitializeWeight}, train for 400 iterations using Adam~\cite{adam} with a learning rate of 0.01 and use 64-dimensional hidden layer and 32-dimensional latent variables in all experiments. 

\subsubsection{Results and Discussions} 
We compare against VGAE \cite{kipf2016variational}, which is the state-of-the-art unsupervised graph convolutional method for link prediction. Spectral clustering (SC)~\cite{Tang2011} and DeepWalk (DW)~\cite{Perozzi2014} are two popular baselines that are also used in the comparison. The results are summarized in Table~\ref{linkprediction_results}. We present the mean and standard error of two metrics, AUC and AP, for 10 runs with random initialization on fixed dataset splits. Our proposed CensNet-VAE has better or similar performance compared to VGAE. Please note, for small citation graphs such as Cora and Citeseer, both VGAE and CensNet-VAE have similar or tied performance. Nevertheless, in large citation graph, Pubmed (which has 3 and 6 times more number of nodes than Cora and Citeseer, respectively), CensNet-VAE achieves significantly higher predictive performance. We can conclude that our node and edge switch co-learning structure obtains better performance than the vanilla GCN, especially in large data set. In terms of the number of layers in the CensNet-VAE method, we evaluated different combinations and numbers of node layers and edge layers, and observed that adding more edge layers will not contribute significant performance gain. This observation is consistent with the current literature, i.e., deeper graph neural networks would perform worse than the shallow ones~\cite{Kipf2016SemiSupervisedCW}. Some recent studies ~\cite{Hasanzadeh2019SIGVAE} also show that using a two-layer GCN as encoder is a common choice for the link prediction task.

Overall, the experimental results show that the proposed CensNet method has competitive and consistent performance over diverse learning tasks on various benchmark datasets.

\section{Conclusions and Future Work}
\label{conclusion}
In this paper, we proposed a novel deep graph learning framework, CensNet, to perform the convolution operations with both node and edge features in graph. Different from existing approaches, CensNet can learn and enhance the node embeddings by incorporating the information from edge features, and it then improves the edge embeddings with node features. This continuously, progressively, and mutually beneficial approach can significantly exploit the information that graph contains. Our model consists of two types of layers, node and edge layers, and such a layer combo composes an iteration of node and edge embedding updates. The CensNet is flexible and powerful for various graph learning tasks including semi-supervised node classification,  graph classification, graph regression, and unsupervised link prediction. We implemented CensNet in these four learning tasks with five benchmark data sets including citation networks and molecular graphs. The experimental results show that the proposed CensNet method can achieve the state-of-the-art performance in most of the tasks. 

For future steps, we are currently exploring other graph kernels and more efficient training algorithms to scale up the deep graph learning to larger and more complex data sets. Specifically, we are interested in the following improvement directions. (1). Can we improve the model performance with more graph layers (i.e., deeper graph neural network)? Current research discovered that the depth can negatively impact on the model performance and it's difficult to train a graph neural network with seven or more layers \cite{Kipf2016SemiSupervisedCW}. We are currently working on new graph kernels and residual connections to avoid such pitfalls. (2). Can we scale up the CensNet architecture to graphs with millions of nodes? Most existing graph neural network models cannot work with such a huge graph, even with mini-batch implementations. We're exploring sampling-based approaches and distributed training algorithms to solve this problem. (3). Dynamic graph neural network. We usually assume a static graph for convolutional operations, while a growing, temporally and spatially, dynamic graph is very common in real work applications. For example,  new nodes and edges (connections) appear in social networks, protein-protein interaction changes over time. We need a scalable, efficient, and dynamic graph neural network to address all of above challenges.


%



\ifCLASSOPTIONcompsoc
  \section*{Acknowledgments}
\else
  \section*{Acknowledgment}
\fi

The authors thank the anonymous reviewers for their helpful comments. This work was supported in part by a gift from Adobe.

\ifCLASSOPTIONcaptionsoff
  \newpage
\fi




\bibliographystyle{IEEEtran}
\bibliography{IEEEabrv,CensNet}
%



%





\begin{IEEEbiography}[{\includegraphics[width=1in,height=1.25in,clip,keepaspectratio]{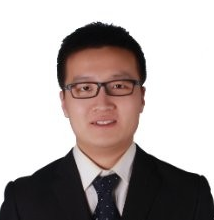}}]{Xiaodong Jiang} received the B.S. in Mathematics from Beijing University of Technology, Beijing, China, in 2014, the M.S. degree in Statistics, Computer Science, and Ph.D. degree in Statistics from the University of Georgia, Athens, GA, in 2016, 2019 and 2019, respectively. He is currently a Research Data Scientist at Facebook Inc., Menlo Park, California. His research areas are graph based machine learning, bioinformatics, and time series analysis. 
\end{IEEEbiography}

\begin{IEEEbiography}[{\includegraphics[width=1in,height=1.25in,clip,keepaspectratio]{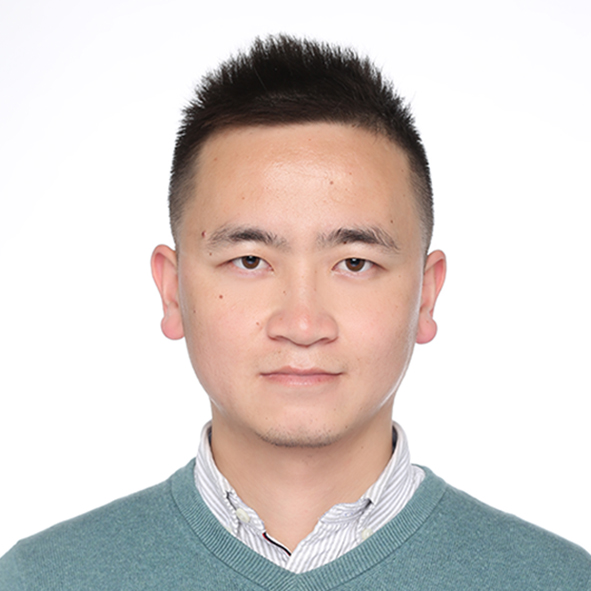}}]{Ronghang Zhu} received the B.S. and M.S. degree in the College of Computer Science from Sichuan University, China, in 2014 and 2017. He is now working towards the Ph.D degree in the Department of Computer Science at University of Georgia. His research interests include domain adaptation and computer vision.
\end{IEEEbiography}

\begin{IEEEbiography}[{\includegraphics[width=1in,height=1.25in,clip,keepaspectratio]{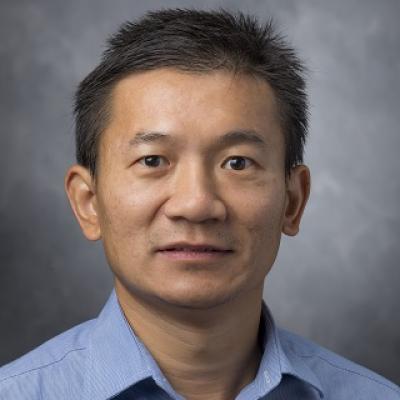}}]{Pengsheng Ji}   received his B.S. degree in mathematics at Nankai University, Tianjin, China, and his Ph.D. degree in statistics in 2012 at Cornell University, Ithaca, NY. He is currently an Associate Professor of statistics at the University of Georgia, and has published in the Annals of Statistics, Annals of Applied Statistics, Journal of Machine Learning Research, International Joint Conference on Artificial Intelligence, Electronic Journal of Statistics, Journal of Social Work, etc.  His research areas are social network analysis, machine learning, bibliometrics, bioinformatics, etc. He received the M. G. Michael Award for his research.
 
\end{IEEEbiography}

\begin{IEEEbiography}[{\includegraphics[width=1in,height=1.25in,clip,keepaspectratio]{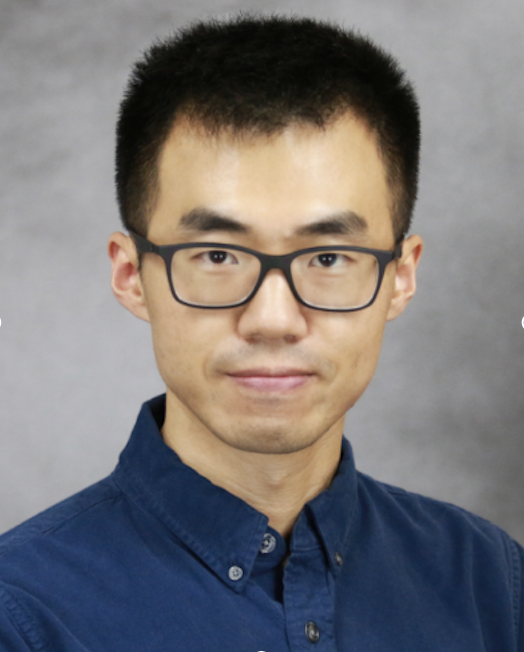}}]{Sheng Li} (S'11-M'17-SM'19) received the B.Eng. degree in computer science and engineering and the M.Eng. degree in information security from Nanjing University of Posts and Telecommunications, China, and the Ph.D. degree in computer engineering from Northeastern University, Boston, MA, in 2010, 2012 and 2017, respectively. He is a Tenure-Track Assistant Professor at the Department of Computer Science, University of Georgia since 2018. He was a research scientist at Adobe Research from 2017 to 2018. He has published over 90 papers at peer-reviewed conferences and journals, and has received over 10 research awards, such as the INNS Aharon Katzir Young Investigator Award, Adobe Data Science Research Award, SDM Best Paper Award, and IEEE FG Best Student Paper Honorable Mention Award. He serves as an Associate Editor of IEEE Computational Intelligence Magazine, Neurocomputing, IET Image Processing and SPIE Journal of Electronic Imaging, and serves on the Editorial Board of Neural Computing and Applications. He has also served as a reviewer for several IEEE Transactions, area chair/senior program committee member for AAAI, IJCAI and ICPR, and program committee member for NeurIPS, ICML, CVPR, ECCV, KDD and ICLR. His research interests include graph based machine learning, representation learning, visual intelligence, user behavior modeling, and causal inference.
\end{IEEEbiography}

\vfill




\end{document}